\documentclass[10pt, a4paper]{article}

\usepackage{lrec-coling2024} 

\usepackage{multirow}

\usepackage[inline]{enumitem}

\usepackage{tabularx,multirow,float}
\newcolumntype{C}{>{\centering\arraybackslash}p{2.5em}}
\usepackage{amsmath,amssymb,booktabs}
\allowdisplaybreaks

\title{Inductive Knowledge Graph Completion with GNNs and Rules:\\ An Analysis}

\name{Akash Anil, V\'ictor Guti\'errez-Basulto, Yazm\'in Ibañ\'ez-Garc\'ia, Steven Schockaert} 

\address{Cardiff University, 
         Cardiff, United Kingdom \\
         {\{anila, gutierrezbasultov, ibanezgarciay, schockaerts1\}}@cardiff.ac.uk\\}

\abstract{
The task of inductive knowledge graph completion requires models to learn inference patterns from a training graph, which can then be used to make predictions on a disjoint test graph. Rule-based methods seem like a natural fit for this task, but in practice they significantly underperform state-of-the-art methods based on Graph Neural Networks (GNNs), such as NBFNet. We hypothesise that the underperformance of rule-based methods is due to two factors: (i) implausible entities are not ranked at all and (ii) only the most informative path is taken into account when determining the confidence in a given link prediction answer. To analyse the impact of these factors, we study a number of variants of a rule-based approach, which are specifically aimed at addressing the aforementioned issues. We find that the resulting models can achieve a performance which is close to that of NBFNet. Crucially, the considered variants only use a small fraction of the evidence that NBFNet relies on, which means that they largely keep the interpretability advantage of rule-based methods. Moreover, we show that a further variant, which does look at the full KG, consistently outperforms NBFNet. 
 \\ \newline  }

\begin{document}

\maketitleabstract

\section{Introduction}
Knowledge Graphs (KGs) store graph-like  factual knowledge using \emph{triples} of the form $(x,r,y)$, indicating that entities $x$ and $y$ are related through a relation type $r$. They provide the backbone of several NLP related tasks, ranging from question answering~\cite{Yu_2019,Zhiwei_2022_2} to recommendation systems~\cite{Yuhao_2022,Ding_2022}.
The aim of Knowledge Graph completion is to find plausible 
triples which are missing from a given KG. 
Standard KG completion models are best suited for densely connected static knowledge graphs. To encourage the study of KG completion methods that can be applied more widely, \citet{DBLP:conf/icml/TeruDH20} introduced the task of \emph{inductive} KG completion (where the usual setting is then referred to as \emph{transductive}). In the inductive setting, we have one knowledge graph for training and a separate knowledge graph for testing. Similar as for the transductive setting, the main evaluation task consists in answering link prediction queries of the form $(x,r,?)$, which require us to identify entities $y$ that make $(x,r,y)$ a valid triple. 
Crucially, in the inductive setting, the entities in the training and test KGs are different, so the model has to make predictions about entities that were not seen during training. We would thus expect that capturing inference patterns becomes crucial, which suggests that rule-based methods should be a natural fit for this setting. However, the literature on inductive KG completion is  dominated by GNN models. 

The aim of this paper is to analyse the reasons why rule-based methods underperform, and to suggest strategies for mitigating the underlying issues. We focus on NBFNet \cite{DBLP:conf/nips/ZhuZXT21}, as a state-of-the-art GNN model, and AnyBURL \cite{DBLP:conf/ijcai/MeilickeCRS19}, as a state-of-the-art rule-based method. We start from the observation that standard  rule-based methods have two key limitations, irrespective of the quality of the learned rules:
\begin{description}
\item[L1 (Zero-confidence entities)] For a link prediction query of the form $(x,r,?)$, rule-based methods only assign a non-zero confidence to entities that can be linked to $x$ by a rule. This means that the majority of entities will typically receive a confidence of 0. Rule-based methods thus make no attempt to rank entities which do not seem plausible.
\item[L2 (Aggregation of evidence)] For most rule-based methods, the confidence in an answer candidate is essentially based on the confidence of a single rule. Intuitively, however, the more (ground) rules predict an answer candidate $y$, the more confident we should be.
\end{description}
We analyse the effect of these issues by comparing the performance of AnyBURL and NBFNet with a hybrid approach. To address limitation L1, we simply rank the entities that receive a confidence of 0 with a different method, which is essentially used as a tie breaker. 
By comparing the performance of this hybrid method with the standard AnyBURL set-up, we can evaluate the impact of ignoring zero-confidence entities on the overall performance. 
To address L2, we essentially want to combine the confidence scores of all the ground rules that predict a given answer candidate. Doing this effectively is challenging, however, because we need to distinguish between cases where different rules provide independent evidence and cases where different variations of the same argument are found \cite{DBLP:conf/akbc/OttMS21,DBLP:conf/esws/BetzMS22}. To address this challenge, we experiment with a GNN-based strategy for aggregating the evidence provided by different rules. 
Since the graphs involved are small, this strategy largely keeps the interpretability advantage of rule-based methods. 


By combining the strategies for addressing L1 and L2, we achieve results which are close to those of NBFNet. As a final variant, we use NBFNet to re-rank the entities predicted by AnyBURL. When also using NBFNet for re-ranking the zero-confidence entities, we end up with a hybrid strategy which consistently outperforms the standard NBFNet, although without the interpretability advantage of rule-based methods. Note that in this variant, AnyBURL is only used to partition to set of candidate answers into those with zero confidence and those that can be predicted by some rule. The fact that this variant outperforms NBFNet suggests that the latter is prone to learning spurious correlations. 


\section{Related Work}

In the last decade, several neural methods for transductive knowledge graph completion (KGC)  have been proposed. Commonly used paradigms for KGC are GNN-based~\cite{DBLP:conf/iclr/VashishthSNT20,DBLP:conf/esws/SchlichtkrullKB18}, rule-based \cite{DBLP:conf/ijcai/MeilickeCRS19,DBLP:conf/iclr/QuCXBT21,DBLP:conf/kr/Wu0WS22} and embedding-based approaches \cite{DBLP:conf/nips/BordesUGWY13,Yankai_2015,TrouillonWRGB_2016,DBLP:conf/iclr/SunDNT19,Shuai_2019,DBLP:conf/emnlp/BalazevicAH19,Zhanqiu_2020,Ines_2020}. A shortcoming of these approaches is that they cannot straightforwardly be used for newly emerging entities. Methods that rely on the textual descriptions of entities and relations~\cite{Liang_2019,Yang_2022,Elan_2022} can overcome this limitation to some extent, but textual descriptions are not always available, which limits their applicability.

Most real-world KGs are dynamic, with new information being continuously added (e.g.\ new items or customers in recommendation systems). It is thus paramount to have methods that can make predictions about new entities, without relying on costly retraining. To be entity-agnostic, methods for inductive KGC have to capture the dependencies that hold between the different relations in the input KG.  Inductive KGC methods currently fall into two main categories: \emph{rule-based methods} and \emph{GNN-based methods}. Classical rule learners, such as AnyBURL \cite{DBLP:conf/ijcai/MeilickeCRS19} and AMIE \cite{DBLP:journals/vldb/GalarragaTHS15}, explore paths in the KG to discover common patterns, and thus induce (weighted) Horn rules~\cite{DBLP:conf/ijcai/OmranWW18,DBLP:conf/aaai/Pirro20}. 
Inspired by these classical methods, differentiable variants, approximating the search of rules, have recently also been studied~\cite{DBLP:conf/nips/YangYC17,DBLP:conf/nips/SadeghianADW19,DBLP:conf/acl/NeelakantanRM15,DBLP:conf/eacl/McCallumNDB17,DBLP:conf/iclr/DasDZVDKSM18,DBLP:conf/iclr/QuCXBT21}. On the other hand, GNNs have become a popular architecture for modelling the topological structure of graphs in different application domains~\cite{DBLP:series/synthesis/2020Hamilton}. Indeed, the inherent capability of GNNs to describe subgraph patterns has led to the development of state-of-the-art GNN-based inductive KGC methods~\cite{DBLP:conf/icml/TeruDH20,DBLP:conf/aaai/MaiZY021,DBLP:conf/nips/ZhuZXT21,DBLP:conf/nips/LiuGHK21,DBLP:conf/icml/YanMGT022,DBLP:conf/www/ZhangY22,DBLP:conf/iclr/0001DWH22}. These methods vary in the way that graph information is used and generated. For example, GraIL \cite{DBLP:conf/icml/TeruDH20} and CoMPILE~\cite{DBLP:conf/aaai/MaiZY021} explicitly encode the subgraph enclosing the  pair of nodes in a triple to predict whether that triple is plausible. Such methods are very expensive as a different subgraph has to be generated for each answer candidate, which is not feasible in large KGs. By contrast, NBFNet \cite{DBLP:conf/nips/ZhuZXT21} and RED-GNN~\cite{DBLP:conf/www/ZhangY22} dynamically generate the set of relational paths between the entity from the query and the different candidate answers. 
Variants of NBFNet which only expand the most promising paths between two nodes have also been investigated~\cite{NBFNetA,DBLP:journals/corr/abs-2205-15319}. 
The CBGNN model \cite{DBLP:conf/www/ZhangY22} uses GNNs to learn the representation of cycles, which is inspired by the close connection between cycles and rules.
RefactorGNN \cite{DBLP:conf/nips/ChenM0MS022} is a GNN-based model for inductive KG completion, which is designed to mimic the gradient descent training dynamic of transductive methods such as DistMult. \cite{DBLP:journals/corr/YangYHGD14a}.
As an alternative to GNNs, NodePiece~\cite{DBLP:conf/iclr/0001DWH22} uses transformers to learn entity representations from a fixed-size vocabulary of tokens. 

To the best of our knowledge, this paper is the first attempt to analyse the performance of rule-based methods for inductive KGC, and to develop strategies to integrate them with GNNs. The closest to our paper is the work by~\citet{DBLP:conf/akbc/MeilickeBS21}, which analyses the performance of AnyBURL in the transductive setting and introduces a hybrid method that combines AnyBURL with KG embeddings. 

%








\section{Background}
We now recall the inductive KG completion setting and provide some background on the two methods we will focus on: AnyBURL and NBFNet.

\smallskip
\noindent \textbf{Inductive KG Completion.}
In inductive KG completion \cite{DBLP:conf/icml/TeruDH20}, we are given two knowledge graphs: $\mathcal{G}_{\textit{train}}$ and $\mathcal{G}_{\textit{test}}$. None of the entities from $\mathcal{G}_{\textit{test}}$ appears in $\mathcal{G}_{\textit{train}}$. However, we are guaranteed that every relation that appears in $\mathcal{G}_{\textit{test}}$ also appears in $\mathcal{G}_{\textit{train}}$. The aim is to learn the parameters $\theta$ of a link prediction model using $\mathcal{G}_{\textit{train}}$ only. Given a link prediction query $(x,r,?)$, this model assigns as score $f_{\theta}(y | x,r,\mathcal{G})$ to candidate answers $y$, reflecting its confidence that $(x,r,y)$ is a valid triple, given the information encoded in $\mathcal{G}$. To train the parameters of the model, for each triple $(x,r,y)\in \mathcal{G}_{\textit{train}}$, one or more corrupted triples $(x,r,y')$ are generated, and the model is trained to score $f_{\theta}(y | x,r,\mathcal{G}_{\textit{train}}\setminus\{(x,r,y)\})$ higher than $f_{\theta}(y' | x,r,\mathcal{G}_{\textit{train}}\setminus\{(x,r,y)\})$, which can be implemented using binary cross-entropy, among others. Once the parameters $\theta$ have been learned, the model is used for link prediction on $\mathcal{G}_{\textit{test}}$. To this end, 10\% of the edges in $\mathcal{G}_{\textit{test}}$ are held out as test triples. 
Given a test triple $(x,r,y)$, the  model is used to score all candidate answers against the query $(x,r,?)$. These scores are then used to rank the candidates. The same process is also used for queries of the form $(?,r,y)$. 

In the experiments reported by \cite{DBLP:conf/icml/TeruDH20}, the set of candidate answers consisted of the answer $y$, along with 50 randomly chosen entities from $\mathcal{G}_{\textit{test}}$. This choice was made because of the limited scalability of their model. In line with standard practice for transductive KG completion, papers on inductive KG completion have recently started to report results for the more natural setting where every entity in $\mathcal{G}_{\textit{test}}$ is considered as a candidate answer \cite{DBLP:conf/www/ZhangY22,DBLP:conf/nips/ChenM0MS022}.

\medskip
\noindent\textbf{AnyBURL}
AnyBURL~\cite{DBLP:conf/ijcai/MeilickeCRS19} has been  designed to work as a rule learner for KGs. 
First note that a triple $(x,r,y)$ can be seen as a first-order atom of the form $r(x,y)$.
A \emph{ground path rule} of length $n$ is a rule of the form $b_1(e_1,e_2) \wedge \dots \wedge b_n(e_n,e_{n+1}) \rightarrow h(e_1,e_{n+1})$, where each $e_i$ is an entity from the KG. Note how the body of such a rule corresponds to a path in the knowledge graph. The rule expresses that when the triples $(e_1,b_1,e_2), \dots, (e_n,b_n, e_{n+1})$ all belong to the KG, we would expect the triple $(e_1,h,e_{n+1})$ to be included as well.
A \emph{closed path rule} is a rule of the form $b_1(X_1,X_2) \wedge \dots \wedge b_n(X_n,X_{n+1}) \rightarrow h(X_1,X_{n+1})$, where the arguments of each relation are now variables. The \emph{groundings} of such a rule are all the ground path rules that can be obtained by replacing all variables with specific entities. The aim of AnyBURL is to learn a set of closed path rules\footnote{AnyBURL can also learn rules with constants. However, as such rules are not relevant for inductive KG completion, we will not consider them here.}, along with their confidence, which intuitively reflects how many of its groundings are satisfied in the given KG. 
Once the rules have been learned, they can be used for answering link prediction queries of the form $(x,r,?)$. The confidence of an answer candidate $y$ is given by the most confident rule which can predict the fact $r(x,y)$. Entities which are not predicted by any of the rules receive a confidence degree of 0. In principle, the answer candidates are ranked according to their confidence degree. However, when a tie between two entities $y_1$ and $y_2$ occurs, their relative position is decided by looking at the second-most confident rule that predicts $r(x,y_1)$ and $r(x,y_2)$, if any. This process can be repeated, by looking at more rules until the tie is broken.

A number of approaches have already been considered for improving AnyBURL by aggregating the confidence scores of different rules that predict the same triple. The most straightforward solution is noisy-or: if a fact is predicted by rules with confidence scores $c_1,...,c_n$ then the overall confidence could be computed as $1-\prod_i (1-c_i)$. However, this ignores the dependencies between the rules and performs poorly in practice \cite{DBLP:conf/akbc/OttMS21}. More recently, \citet {DBLP:conf/esws/BetzMS22} proposed a strategy where each rule $r$ is represented by an embedding $\mathbf{r}\in \mathbb{R}^d$. The overall confidence is then obtained by max-pooling the embeddings of the rules that predict a given fact and applying noisy-or to the coordinates of the resulting vector. This strategy was found to outperform AnyBURL, but the improvements were small for standard KGs.

\medskip
\noindent\textbf{NBFNet}
%
Given a query of the form $(x,r,?)$, 
NBFNet ~\cite{DBLP:conf/nips/ZhuZXT21} learns an embedding $\mathbf{h_{x,y}}$ for each entity $y$ in the knowledge graph. The underlying intuition is to compute the \emph{generalised sum} ($\oplus$) of the representations of all paths between $x$ and $y$, with each path representation being computed as the \emph{generalised product} ($\otimes$) of the representations of the edges in the path. The approach takes inspiration of the Bellman-Ford algorithm for computing the shortest paths from a node in graph to all the other nodes. Crucially, for a given head entity $x$ and a given query relation $r$, NBFNet computes the embeddings $\mathbf{h_{x,y}}$ for all entities $y$ in the graph in a single pass. This makes it significantly more scalable than methods such as GraIL \cite{DBLP:conf/icml/TeruDH20}, which construct a different graph for each possible entity $y$ and need to apply a GNN model to each of these graphs. 
Note that despite the intuitive link with Bellman-Ford, the embeddings $\mathbf{h_{x,y}}$ can depend on nodes which are not part of any path between $x$ and $y$ (let alone the shortest path).
\section{Hybrid Link Prediction Strategies}\label{secHybridLinks}
Our focus is on analysing Limitations L1 and L2 from the introduction. In this section, we propose GNN-based strategies to specifically address each of these limitations. 
For a query $q=(x,r,?)$, we write $\mathcal{A}_q$ for the set of entities which are predicted by AnyBURL with non-zero confidence, and $\mathcal{B}_q$ for the remaining set of entities. Let us consider the Hits@k evaluation metric, i.e.\ we are interested in whether the correct answer $y$ is among the $k$ highest ranked entities. With AnyBURL, there are three cases where this may not be the case:
\begin{enumerate}
\item We have $y\in \mathcal{B}_q$ and $|\mathcal{A}_q|< k$. In this case, the error comes from the fact that the entities in $\mathcal{B}_q$ are not ranked when using AnyBURL.
\item We have $y\in \mathcal{B}_q$ and $|\mathcal{A}_q|\geq k$. In other words, AnyBURL predicts at least $k$ plausible entities, but none of them are the correct answer.
\item We have $y\in \mathcal{A}_q$, but there are at least $k$ other entities in $\mathcal{A}_q$ which are ranked higher than $y$. In this case, AnyBURL identifies $y$ as a plausible candidate, but with insufficient confidence.
\end{enumerate}
To address Case 1, which corresponds to Limitation L1, we need to introduce a strategy for ranking the entities in $\mathcal{B}_q$. Since these entities are deemed to be implausible by AnyBURL, we need a strategy which does not rely on rule-like inference patterns. 
We will use NBFNet for ranking the entities in $\mathcal{B}_q$, as this will facilitate the analysis. 
Case 2 covers a variety of failures, including situations where the KG simply does not provide sufficient evidence for identifying the correct answer $y$. We are particularly interested in situations where NBFNet correctly predicts $y$ among the top-$k$ entities. In our analysis, we found that such situations are in fact rare.
We therefore focus on Case 3 in the remainder of this section. 

\subsection{Reranking with GNNs}\label{secRerankingGNN}
Many of the rules which are learned by AnyBURL are soft rules, in the sense that they do not \emph{necessarily} imply that the predicted relation is valid, but rather provide some degree of evidence in support of that prediction. The number of different rules that predict a given answer candidate thus matters. The problem of rule aggregation was already addressed by \citet{DBLP:conf/esws/BetzMS22}, but their approach only looks at whether two rules capture a similar argument. For instance, if we want to predict whether $X$ speaks language $Z$, then $\textit{livesIn}(X,Y) \wedge \textit{hasLanguage}(Y,Z)\rightarrow \textit{speaks}(X,Z)$ captures a similar argument to $\textit{worksIn}(X,Y) \wedge \textit{hasLanguage}(Y,Z)\rightarrow \textit{speaks}(X,Z)$. Knowing that both rules are satisfied therefore does not add much evidence compared to only knowing that one of the rules is satisfied. However, they do not take into account how many groundings of each rule are satisfied, which can be important. 
Moreover, they do not take into account whether different ground rules rely on the same triples or not, while this can also be important for determining whether rules provide independent evidence. Finally, and perhaps most crucially, the latent rule embedding approach \cite{DBLP:conf/esws/BetzMS22} requires sufficient training data to learn an embedding of each rule. In our experiments, we did not manage to achieve non-trivial results with this method, due to the small size of the training graphs in the inductive KG completion benchmarks. To address these issues, we propose a strategy which aggregates AnyBURL rules using a GNN.


\smallskip
\noindent\textbf{Rule Instantiation Graphs.}
Each rule that predicts the entity $y$ corresponds to a path in the KG from the head entity $x$ to $y$. The set of rules that predict $y$ can thus be viewed as a graph, which we call the \emph{rule instantiation graph}. More formally, let $\mathcal{G}$ be the given knowledge graph. Let $\rho_1,...,\rho_k$ be the ground rules that predict some answer candidate $y$. Each rule $\rho_i$ has the following form:
$$
r_{(i,0)}(x,y_{(i,1)}) \wedge ... \wedge r_{(i,n_i)}(y_{(i,n_i)},y) \rightarrow r(x,y)
$$
Since $\rho_i$ allows us to predict the answer candidate $y$, the body of the rule must be satisfied in $\mathcal{G}$. In other words, the corresponding triples must be included in $\mathcal{G}$. The rule instantiation graph is defined as the union of these triples for the different rules $\rho_1,...,\rho_k$. Note how rule instantiation graphs thus summarise which rules apply, and how the evidence that is used by these rules overlaps.

When a given answer is predicted by a large number of ground rules, we only include the most confident rules when creating the rule instantiation graph. By doing this, we ensure that the rule instantiation graph is focused on the most reliable evidence. Specifically, we limit the rule instantiation graphs to the five ground rules with the highest confidence.\footnote{In the case of ties, we include all ground rules with the same confidence degree as the fifth most confident ground rule.} Another reason to consider only a subset of the rules is to keep our method efficient. Since we need to run the GNN for each entity in $\mathcal{A}_q$, it is important that the rule instantiation graphs remain sufficiently small. For instance, we want to ensure that the rule instantiation graphs for all entities in $\mathcal{A}_q$ can be evaluated in the same mini-batch.


\smallskip
\noindent\textbf{Confidence Estimation.}
We use a GNN to predict the plausibility of {an answer candidate $y$ from their corresponding} rule instantiation graphs. Compared to standard GNN models for link prediction, such as GraIL \cite{DBLP:conf/icml/TeruDH20}, this approach has a number of important advantages. First, we only apply this strategy to the entities in $\mathcal{A}_q$, which is typically much smaller than the full set of entities. Rule instantiation graphs are also much smaller than the graphs that are normally considered ({which typically contain} all $k$-hop paths between the head and tail entity). Our strategy thus avoids the computational inefficiency of GraIL. Furthermore, rule instantiation graphs are focused on evidence that was already uncovered by AnyBURL. In our case, the GNN thus merely needs to learn how to aggregate this evidence. In contrast, in standard approaches, the GNN is presented with a large graph, where only a small fragment of that graph is actually relevant. Our hypothesis is that GNNs are more likely to learn spurious patterns in this standard setting. 

While various types of GNNs can be used for predicting confidence degrees from rule instantiation graphs, we will experiment with the R-GCN \cite{DBLP:conf/esws/SchlichtkrullKB18} and CompGCN \cite{DBLP:conf/iclr/VashishthSNT20} architectures. We include R-GCN because it is one of the most commonly used models for multi-relational graphs. However, R-GCNs are prone to overfitting and are relatively inefficient. For this reason, we also include results for CompGCN, which addresses these issues. In both cases, we compute the initial node embeddings in the same way as GraIL. In particular, for a given node $n$ we compute its distance to the head entity $x$ and the tail entity $y$ (where we identify nodes with the entities they refer to for the ease of presentation). As the input embedding $\mathbf{n^{(0)}}$, we then simply concatenate the one-hot encoding of both distances. These node embeddings are then updated using standard R-GCN or CompGCN layers.
We assume that for each triple $(x,r,y)$ a corresponding inverse triple $(y,r^{-1},x)$ is also included. To prevent overfitting, for the R-GCN model, we apply basis-decomposition \cite{DBLP:conf/esws/SchlichtkrullKB18}. All weight matrices are then learned as a linear combination of $\ell$ base matrices.
The confidence in the answer candidate $y$ for a link prediction query of the form $q=(x,r,?)$ is computed as follows:
\begin{align*}
\textit{conf}(y;q) = \sigma\left(\mathbf{A} (\mathbf{y}\oplus \mathbf{r}) + b\right)
\end{align*}
Here $\mathbf{r}$ is a learned embedding for relation $r$, $\mathbf{y}$ is the embedding of $y$ in the final layer of the GNN, $\oplus$ denotes vector concatenation, $\mathbf{A}$ is a learned linear transformation, and $b$ is a bias term. The model is trained using binary cross-entropy. 

\subsection{Reranking with NBFNet}\label{secNBFNBF}
As already mentioned, we can use NBFNet for ranking the entities in $\mathcal{B}_q$ to address Limitation L1. As a simple hybrid strategy, we will evaluate a method which also uses NBFNet for reranking the entities in $\mathcal{A}_q$. We will refer to this method as \emph{NBFNet + NBFNet}, since it relies on the application of NBFNet to two disjoint sets of entities (i.e.\ $\mathcal{A}_q$ and $\mathcal{B}_q$). Different from the method based on rule instantiation graphs, NBFNet + NBFNet can take into account all the available evidence. By comparing this strategy with the strategy based on rule instantiation graphs, we will thus be able to assess to what extent important information is missing from the rule instantiation graph. By comparing the NBFNet + NBFNet method with the standard NBFNet, we will be able to assess to what extent AnyBURL is capable of identifying the most plausible entities.

\begin{figure}
\centering
\includegraphics[width=190pt, height=130pt]{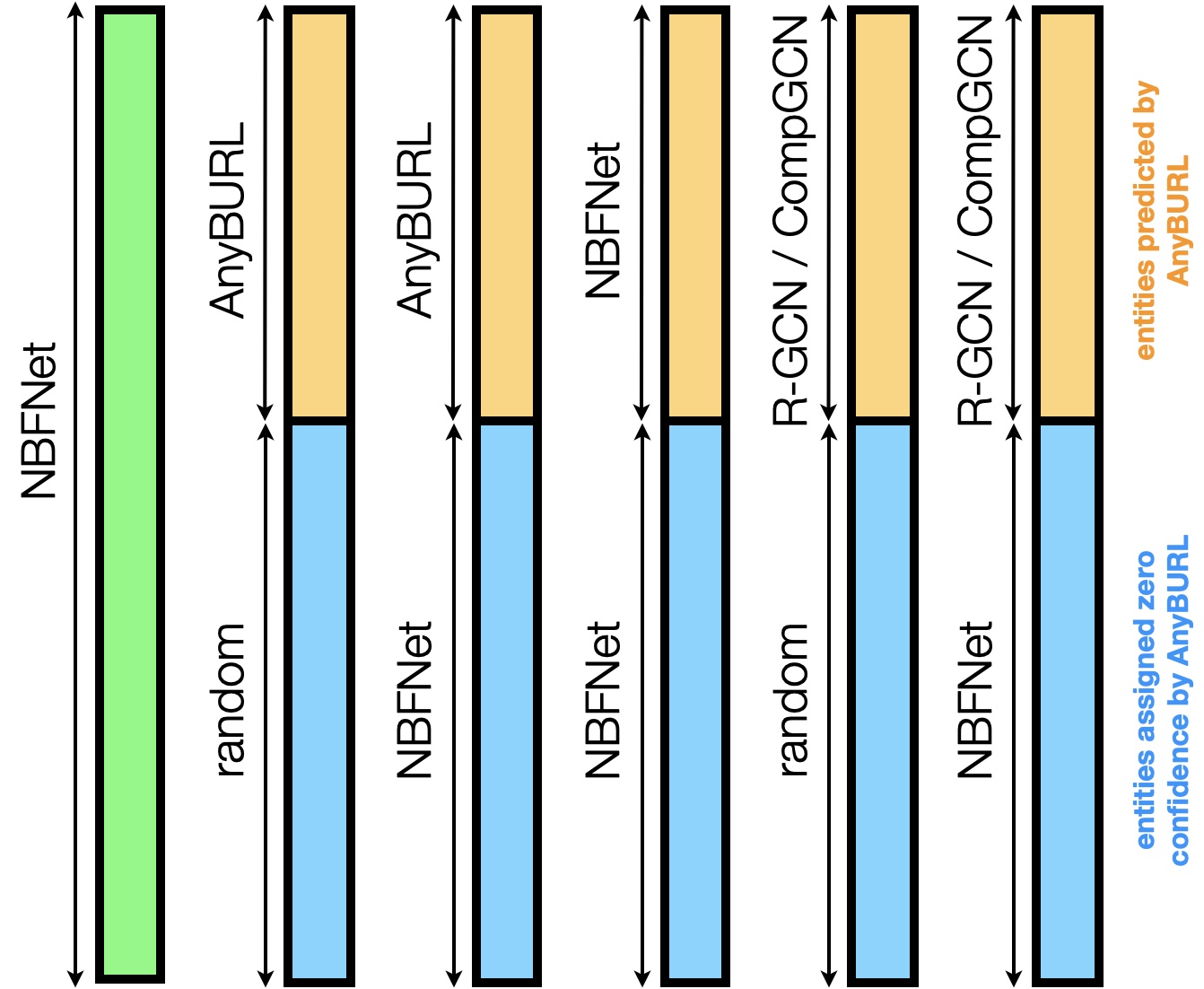}
\caption{Considered hybrid strategies.
\label{figSchematicOverview}}
\end{figure}

\begin{table*}
\centering
\setlength{\tabcolsep}{3pt}
\caption{Overview of the results for the full evaluation setting, where all entities are ranked, on FB15k-237. Results are reported in terms of Mean Reciprocal Rank, Hits@1, Hits@3 and Hits@10.\label{tabMainFB}}

\resizebox{\textwidth}{!}{ \begin{tabular}{l CCCCCCCCCCCCCCCC}
\toprule
& \multicolumn{16}{c}{\textbf{FB15k-237}}\\
& \multicolumn{4}{c}{\textbf{v1}} &\multicolumn{4}{c}{\textbf{v2}} & \multicolumn{4}{c}{\textbf{v3}} & \multicolumn{4}{c}{\textbf{v4}} \\
\cmidrule(lr){2-5}\cmidrule(lr){6-9}\cmidrule(lr){10-13}\cmidrule(lr){14-17}
& MRR & H@1 & H@3 & H@10 & MRR & H@1 & H@3 & H@10 & MRR & H@1 & H@3 & H@10 & MRR & H@1 & H@3 & H@10 \\
\midrule   
NBFNet           & 45.4 &	35.1 &	\textbf{52.7} &	61.6 &	50.0 &	38.3 &	57.7 &	69.6 &	47.5 &	37.3 &	54.1 &	65.3 &	44.7 &	33.5 &	51.8 &	65.4 \\
AnyBURL          & 35.5 &	30.1 &	39.3 &	43.6 &	43.7 &	33.7 &	49.8 &	60.9 &	33.8 &	25.5 &	37.7 &	49.6 &	31.8 &	23.3 &	35.8 &	48.1\\
\midrule
Noisy-or  & 34.2 &	27.3 &	39.7 &	46.1 &	39.9 &	28.5 &45.5 &	62.8 & 38.3 & 27.7 & 43.8 &	58.6 & 34.8 & 24.6 & 39.3 & 55.0 \\
R-GCN            & 35.7 &	29.3 &	40.4 &	45.3 &	42.7 &	32.7 &	49.8 &	59.5 &	39.8 &	30.7 &	46.0 &	55.4 &	36.4 &	27.4 &	42.0 &	52.3 \\
CompGCN  & 37.0 &	31.4 &	41.3 & 45.6	 & 45.1	& 36.2	& 50.3	& 60.4		& 40.7 & 31.5	& 47.3	& 55.8	& 38.3 & 29.5 & 43.4 & 53.3\\
\midrule
AnyBURL + NBFNet & 42.9 &	33.3 &	48.9 &	59.0 &	47.3 &	35.8 &	54.2 &	67.0 &	36.5 &	27.2 &	40.7 &	53.8 &	34.7 &	25.4 &	38.9 &	52.8\\
Noisy-or + NBFNet  & 41.6 &	30.5 &	49.3 &	61.8 &	43.4 & 30.6 &	49.9 &	68.9 &	40.9 &	29.4 &	46.7 &	62.8 & 37.7 & 26.7 & 42.5 & 59.7 \\
R-GCN + NBFNet   & 43.1 &	32.5 &	50.1 &	61.0 &	46.2 &	34.8 &	54.1 &	65.6 &	42.4 &	32.4 &	48.9 &	59.6 &	39.3 &	29.5 &	45.2 &	57.0\\
CompGCN + NBFNet & 44.4 &	34.5 &	50.8 &	61.1 & 48.7	& 38.3 & 54.6 &	66.6 & 43.3 & 33.2 &	50.2 &	60.0 & 41.1 & 31.6 & 46.6 &	58.0\\
\midrule
NBFNet + NBFNet  & \textbf{46.8} &	\textbf{37.7} &	52.4 &	\textbf{61.9} &	\textbf{52.1} &	\textbf{40.9} &	\textbf{59.3} &	\textbf{70.7} &	\textbf{49.2} &	\textbf{39.3} &	\textbf{55.7} &	\textbf{66.3} &	\textbf{46.9} &	\textbf{36.0} &	\textbf{53.6} &	\textbf{66.4}\\
\bottomrule
\end{tabular}}
\end{table*}

\begin{table*}
\centering
\setlength{\tabcolsep}{3pt}
\caption{Overview of the results for the full evaluation setting, where all entities are ranked, on WN18RR. Results are reported in terms of Mean Reciprocal Rank, Hits@1, Hits@3 and Hits@10.\label{tabMainWN}}
\resizebox{\textwidth}{!}{ \begin{tabular}{l CCCCCCCCCCCCCCCC}
\toprule
& \multicolumn{16}{c}{\textbf{WN18RR}}\\
& \multicolumn{4}{c}{\textbf{v1}} &\multicolumn{4}{c}{\textbf{v2}} & \multicolumn{4}{c}{\textbf{v3}} & \multicolumn{4}{c}{\textbf{v4}} \\
\cmidrule(lr){2-5}\cmidrule(lr){6-9}\cmidrule(lr){10-13}\cmidrule(lr){14-17}
& MRR & H@1 & H@3 & H@10 & MRR & H@1 & H@3 & H@10 & MRR & H@1 & H@3 & H@10 & MRR & H@1 & H@3 & H@10 \\
\midrule
NBFNet & 70.2 &	61.7 &	77.2 &	\textbf{83.2} & 66.3 & 57.7 & 73.4 &	79.2 & 42.7 & 35.3 & 46.9 &	55.0 & 60.0 &	52.7 &	65.6 &	71.0\\
AnyBURL & 39.1 &	23.3 &	44.8 &	76.2 &	54.8 &	41.7 &	66.1 &	77.4 &	34.0 &	26.4 &	39.4 &	46.0 &	55.0 &	45.8 &	62.0 &	70.5 \\
\midrule
Noisy-or  & 57.3 &	46.0 &	65.6 &	77.4 &	63.2 &	55.0 & 70.3 &	77.2 &	39.7 &	34.7 &	42.3 &	48.6 & 54.6 & 45.7 & 61.2 & 71.2 \\
R-GCN & 68.1 &	60.7 &	73.8 &	80.3 & 66.1 &	60.3 &	70.0 &	76.4 & 40.9 &	36.8 &	43.0 &	48.1 & 60.7 &	55.2 &	64.3 &	70.5\\
CompGCN   & 68.1 &	62.0 &	72.1 &	78.4 & 65.3 &	59.0 &	69.9 &	76.3 & 40.7 &	36.0 &	43.4 &	48.5 & 60.3 &	54.5 &	64.1 & 70.1\\
\midrule
AnyBURL + NBFNet & 39.5 &	23.4 &	44.9 &	76.9 &	55.4 &	41.9 &	66.4 &	79.0 &	36.0 &	27.1 &	41.2 &	50.4 &	55.3 &	45.9 &	62.2 &	71.0\\
Noisy-or + NBFNet  & 57.7 &	46.1 &	65.7 &	78.2 &	63.8 &	55.1 &	70.7 &	78.8 &	41.7 &	35.4 &	44.1 &	53.1 & 54.9 & 45.7 & 61.4 & 71.7 \\
RGCN + NBFNet & 68.5 &	60.7 &	74.0 &	81.1 & 66.7 &	60.5 &	70.3 &	77.9 & 42.9 &	37.5 &	44.8 &	52.6 &  61.0 &	55.2 &	64.5 &	71.0\\
CompGCN + NBFNet   & 68.5 &	62.0 &	72.3 &	79.1 & 65.9 & 59.2 & 70.2 &	77.9 & 42.6 & 36.7 & 45.1 &	52.8 & 60.6 & 54.6 & 64.3 &	70.6 \\
\midrule
NBFNet + NBFNet & \textbf{72.5} & \textbf{65.8} &	\textbf{77.4} &	82.2 & \textbf{68.7} & \textbf{61.4} & \textbf{74.4} &	\textbf{79.4} & \textbf{44.8} &	\textbf{38.5} &	\textbf{47.9} &	\textbf{56.1} & \textbf{63.2} &	\textbf{57.0} &	\textbf{68.0} &	\textbf{73.0}\\
\bottomrule
\end{tabular}}
\end{table*}

\begin{table*}
\centering
\setlength{\tabcolsep}{3pt}
\caption{Overview of the results for the full evaluation setting, where all entities are ranked, on NELL-995. Results are reported in terms of Mean Reciprocal Rank, Hits@1, Hits@3 and Hits@10.\label{tabMainNELL}}
\resizebox{\textwidth}{!}{ \begin{tabular}{l CCCCCCCCCCCCCCCC}
\toprule
& \multicolumn{16}{c}{\textbf{NELL-995}}\\
& \multicolumn{4}{c}{\textbf{v1}} &\multicolumn{4}{c}{\textbf{v2}} & \multicolumn{4}{c}{\textbf{v3}} & \multicolumn{4}{c}{\textbf{v4}} \\
\cmidrule(lr){2-5}\cmidrule(lr){6-9}\cmidrule(lr){10-13}\cmidrule(lr){14-17}
& MRR & H@1 & H@3 & H@10 & MRR & H@1 & H@3 & H@10 & MRR & H@1 & H@3 & H@10 & MRR & H@1 & H@3 & H@10 \\
\midrule
NBFNet           & 52.8 &	48.8 &	52.5 &	58.5 &	43.7 &	31.7 &	50.4 &	66.9 &	45.2 &	34.7 &	50.4 &	65.9 &	34.5 &	23.3 &	39.4 &	58.1 \\
AnyBURL          & 47.7 &	38.6 &	49.9 &	61.4 &	43.1 &	33.0 &	48.4 &	61.2 & 39.1 &	30.7 &	43.3 &	55.0 & 33.2 &	23.6 &	39.8 &	51.3\\
\midrule
Noisy-or  & 53.5 &	47.0 &	54.2 &	64.2 &	39.6 &	29.0 &	44.9 &	60.8 &	31.0 &	23.4 &	34.5 &	44.7 & 33.8 & 24.1 & 40.9 & 52.7 \\
R-GCN            & 50.8 &	39.2 &	57.1 &	65.9 &	38.5 &	28.6 &	43.9 &	55.8 &	32.1 &	23.2 &	35.5 &	49.6 &	33.9 &	24.6 &	40.2 &	51.4\\
CompGCN  &48.7 &	38.5 &	53.6 &	63.5 & 40.3 &	30.9 &	47.1 &	55.2 & 34.7 &	25.3 &	39.9 &	52.6 & 33.3 &	23.6 &	40.0 &	51.3\\ 
\midrule
AnyBURL + NBFNet & 48.2 &	39.1 &	50.4 &	61.8 &	46.0 &	34.4 &	52.4 &	67.5 & 45.5 &	35.5 &	50.3 &	64.5 & 38.4 & 26.7 & 45.3 & 60.2 \\
Noisy-or + NBFNet  & 54.0 &	47.5 &	54.7 &	64.6 &	42.7 &	30.5 &	48.9 &	67.1 &	37.4 &	28.2 &	41.5 &	54.2 &	39.3 &	27.1 &	46.3 & 61.6 \\
R-GCN + NBFNet   & 51.2 &	39.7 &	\textbf{57.6} &	66.4 &	41.3 &	30.0 &	47.9 &	62.1 &	38.4 &	28.0 &	42.5 &	59.0 &	39.2 &	27.7 &	45.6 &	60.3 \\
CompGCN + NBFNet &   49.2	 & 39.0 & 54.1 & 64.0 & 43.5 &	32.4 &	51.1 &	61.5 & 41.2 & 30.1 &	46.9 & 62.1 & 38.5 & 26.7 &	45.4 & 60.2\\
\midrule
NBFNet + NBFNet  & \textbf{56.7} &	\textbf{50.8} &	55.7 &	\textbf{68.2} &	\textbf{50.1} &	\textbf{37.8} &	\textbf{57.4} &	\textbf{72.3} &	\textbf{48.3} &	\textbf{37.7} &	\textbf{53.6} &	\textbf{68.6} &	\textbf{39.9} &	\textbf{28.0} &	\textbf{47.2} &	\textbf{62.2}\\
\bottomrule
\end{tabular}}
\end{table*}

\begin{table*}
\centering
\setlength{\tabcolsep}{3pt}
\caption{Overview of the results for the reduced evaluation setting, where only 50 negatives are considered for each query. Results are reported in terms of Hits@10. \label{tabHITS50}}
\resizebox{0.85\textwidth}{!}{\begin{tabular}{l CCCC CCCC CCCC }
\toprule
& \multicolumn{4}{c}{\textbf{FB15k-237}} & \multicolumn{4}{c}{\textbf{WN18RR}} & \multicolumn{4}{c}{\textbf{NELL-995}}\\
\cmidrule(lr){2-5}\cmidrule(lr){6-9}\cmidrule(lr){10-13}
& \textbf{v1} & \textbf{v2} & \textbf{v3} & \textbf{v4} & \textbf{v1} & \textbf{v2} & \textbf{v3} & \textbf{v4} & \textbf{v1} & \textbf{v2} & \textbf{v3} & \textbf{v4}\\ 
\midrule
GraIL \cite{DBLP:conf/icml/TeruDH20} &  64.2&  81.8 &  82.8& 89.3  & 82.5 & 78.7 & 58.4 & 73.4 & 59.5 & 93.3 & 91.4 & 73.2\\
CoMPILE \cite{DBLP:conf/aaai/MaiZY021}& 67.6 & 82.9 & 84.6 & 87.4& 83.6 &79.8 &60.6 &75.4 & 58.3 & 93.8 &92.7 & 75.1\\
NBFNet \cite{DBLP:conf/nips/ZhuZXT21} &  83.4 &  94.9&  95.1&  96.0 & 94.8 & 90.5 & 89.3 & 89.0 & - & - & - & -\\
\midrule
NBFNet  & \textbf{84.5} & 94.9 & 94.6 & 94.7 & 94.6 & 89.7 & 90.4 & 88.9 & 64.4&	95.3&	96.7 &	92.8\\
AnyBURL & 60.4 &	82.3 &	84.7 &	84.9 &	86.7 &	82.8 &	65.6 &	79.6 & 68.3&	83.5&	79.8&	65.2\\
\midrule
Noisy-or  & 59.9 & 82.2 & 84.9 & 85.2 & 86.5 & 82.6 & 66.5 & 79.8 & 71.8 & 83.7 & 80.1 & 65.7 \\
R-GCN & 61.0 & 82.4 & 82.6 & 84.2 & 86.9 & 82.6 & 65.0 & 79.9 & 71.7&	83.5&	79.8&	65.2\\
CompGCN & 60.4 & 82.0 & 83.1 & 84.7 & 86.4 & 82.5 & 65.5 & 79.7 & 67.9 & 83.3 & 80.0 & 65.8 \\
\midrule
AnyBURL + NBFNet & \textbf{84.5} &	\textbf{95.4} &	95.2 &	95.6 &	\textbf{94.9} &	89.8 &	\textbf{90.5} &	\textbf{89.1} & 68.7&	\textbf{96.2}&	\textbf{97.3}&	92.5\\
Noisy-or + NBFNet  & \textbf{84.5} & \textbf{95.4} &	\textbf{95.3} &	\textbf{95.7} &	\textbf{94.9} &	89.8 &	\textbf{90.5} &	\textbf{89.1} &	72.2 &	\textbf{96.2} &	\textbf{97.3} &	\textbf{92.9} \\
R-GCN + NBFNet & \textbf{84.5} & \textbf{95.4} & 93.3 & 94.7 & \textbf{94.9} & 89.8 & \textbf{90.5} & \textbf{89.1} & 72.1&	\textbf{96.2}&	\textbf{97.3}&	92.5\\
CompGCN + NBFNet & \textbf{84.5} & \textbf{95.4} & 93.6 & 95.0 & \textbf{94.9} & \textbf{89.9} & \textbf{90.5} & \textbf{89.1} & 68.3 & \textbf{96.2} & \textbf{97.3} & 92.5 \\
\midrule
NBFNet + NBFNet & \textbf{84.5} & \textbf{95.4} & \textbf{95.3} & 95.6 & \textbf{94.9} & 89.8 & \textbf{90.5} & \textbf{89.1} & \textbf{79.0} &	\textbf{96.2}&	\textbf{97.3}&	\textbf{92.9}\\
\bottomrule
\end{tabular}}
\end{table*}

\section{Experiments}\label{secExperiments}
We now present our experimental analysis\footnote{\href{https://github.com/anilakash/IndKGC}{https://github.com/anilakash/IndKGC}}.

\smallskip
\noindent\textbf{Datasets.}
We use the benchmarks for inductive KG completion that were introduced by \citet{DBLP:conf/icml/TeruDH20}. These benchmarks were derived from three well-known benchmarks for transductive KG completion: FB15k-237, WN18RR and NELL-995. Specifically, starting from a given knowledge graph, they randomly select a number of entities and take their $k$-hop neighbourhoods. The resulting subgraph is used as the training graph $\mathcal{G}_{\textit{train}}$. Then they remove the entities from this training graph and repeat the same process to sample a disjoint test graph $\mathcal{G}_{\textit{test}}$. 
Four variants were obtained for each of the three knowledge graphs. For each of the resulting graphs, 10\% of the triples  are removed as a validation split and another 10\% of the edges are removed as a test split. For each of the 12 datasets, we thus have 6 graphs: the training, validation and test splits of $\mathcal{G}_{\textit{train}}$, and the training, validation and test splits of $\mathcal{G}_{\textit{test}}$.




\smallskip
\noindent\textbf{Experimental Setup}
In the experiments by \citet{DBLP:conf/icml/TeruDH20}, the test split of $\mathcal{G}_{\textit{train}}$ and the validation split of $\mathcal{G}_{\textit{test}}$ are not used: the models are trained on the training split of $\mathcal{G}_{\textit{train}}$ and the validation split of $\mathcal{G}_{\textit{train}}$ is used for tuning. The trained models are then tested on the test split of $\mathcal{G}_{\textit{test}}$, using the training split of $\mathcal{G}_{\textit{test}}$ as the fact graph. We follow the same approach, which was also adopted for NBFNet \cite{DBLP:conf/nips/ZhuZXT21}. We noticed that for several other methods, including RED-GNN~\cite{DBLP:conf/www/ZhangY22}, CBGNN~\cite{DBLP:conf/icml/YanMGT022}, NodePiece \cite{DBLP:conf/iclr/0001DWH22} and RefactorGNN \cite{DBLP:conf/nips/ChenM0MS022}, a different methodology was used, which means that their published results are not comparable to the numbers we report in this paper. In particular, \citet{DBLP:conf/www/ZhangY22} use the test split of $\mathcal{G}_{\textit{train}}$ during training (as a validation set) and evaluate on both the validation and test splits of $\mathcal{G}_{\textit{test}}$; \citet{DBLP:conf/icml/YanMGT022} use the training split of $\mathcal{G}_{\textit{test}}$ for validation; and \citet{DBLP:conf/iclr/0001DWH22} use the validation split of $\mathcal{G}_{\textit{test}}$ for validation. For RefactorGNN, we were not able to determine the methodology that was used based on the publicly available code.  
Another difference in the literature concerns the negative examples in the test set. \citet{DBLP:conf/icml/TeruDH20} only considered 50 randomly sampled entities as negative examples, whereas more recent papers considered the setting where all entities from the KG are candidates. We will consider both settings.
%

\smallskip
\noindent\textbf{Strategies} 
Our focus is on comparing the proposed hybrid strategies with \textbf{AnyBURL} and \textbf{NBFNet}. As alternatives to AnyBURL, we use the confidence scores of the \textbf{R-GCN} or \textbf{CompGCN} model for ranking the entities in $\mathcal{A}_q$. We also experiment with \textbf{Noisy-or} as a rule aggregation strategy. We tried to use the sparse \emph{Rule embedding} strategy from \cite{DBLP:conf/esws/BetzMS22} but it could not learn anything meaningful because of the size of the training set. For AnyBURL and the aforementioned variants, only the entities from $\mathcal{A}_q$ are ranked. To allow for a fair comparison, we append the entities from $\mathcal{B}_q$ at the bottom of the ranking, in a randomly shuffled order. Finally, we consider a number of variants in which the entities in $\mathcal{B}_q$ are ranked with NBFNet. For instance, we write \textbf{AnyBURL + NBFNet} for the strategy where the entities from $\mathcal{A}_q$ are ranked by their AnyBURL confidence, and then the entities from $\mathcal{B}_q$ are appended, ranked according to their NBFNet confidence. In the same way, we consider \textbf{Noisy-or + NBFNet}, \textbf{R-GCN + NBFNet}, \textbf{CompGCN + NBFNet} and \textbf{NBFNet + NBFNet}. Figure \ref{figSchematicOverview} provides a schematic overview of these different strategies. 
Given the aforementioned methodological differences, we cannot compare with the published results of many baselines. Furthermore, the NBFNet paper only reported results for the setting with 50 randomly sampled negatives, so we have re-evaluated that model on the full setting. For the setting with 50 randomly sampled negatives, we also compare with the published results of GraIL \cite{DBLP:conf/icml/TeruDH20} and CoMPILE \cite{DBLP:conf/aaai/MaiZY021}, which use the same setting.

\begin{figure*}
\centering
\includegraphics[width=470pt]{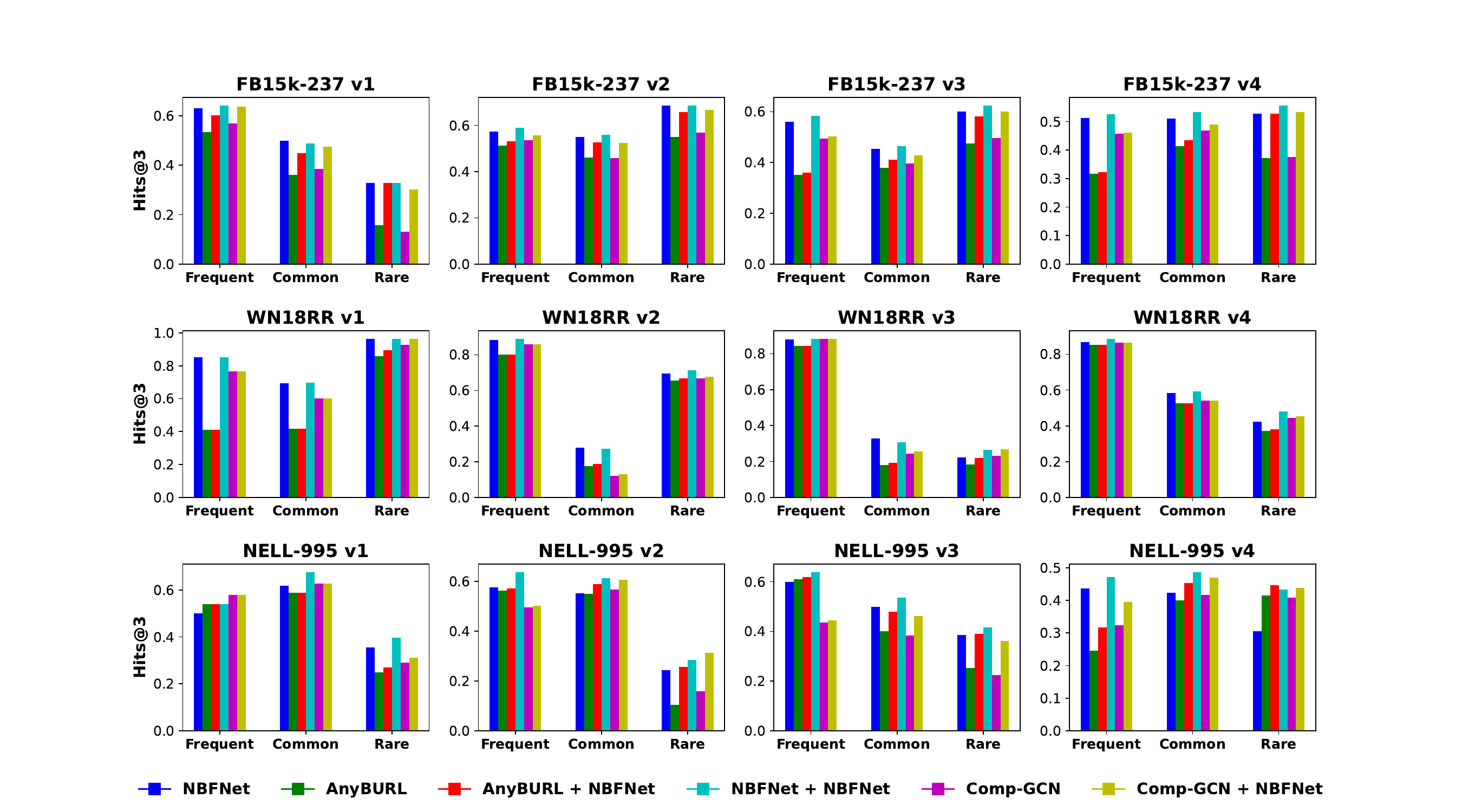}
\caption{Hits@3 performance for frequent (10\% most frequent), common (10-50\% most frequent) and rare (50\% least frequent) relations.\label{figCommonRare}}
\end{figure*}


\smallskip
\noindent
To train the R-GCN and CompGCN models, we iterate over all the triples in the training split of $\mathcal{G}_{\textit{train}}$. Each triple $(x,r,y)$ is used to create two link prediction queries, namely $(x,r,?)$ and $(?,r,y)$. For each query, 
we also need one or more negative examples. Randomly chosen negative examples often have an empty rule instantiation graph, which makes them unsuitable for training. Therefore, we select two random negative examples but with a non-empty rule instantiation graph. 
%
%
AnyBURL is an anytime algorithm, which continues to learn rules for a fixed amount of time. For our experiments, we set this time limit to 10 seconds. We learn rules with up to four body atoms. For the R-GCN, we set the size of the basis $\ell$ to 4 and the number of layers to 4. The model is trained using Adam with an initial learning rate of 0.004 and early stopping patience of 3. For the CompGCN model, we used a learning rate of 0.001 and otherwise used the same values as for the R-GCN. 
We rely on the original implementations of AnyBURL (https://tinyurl.com/yhfd96dv) 
and NBFNet (https://tinyurl.com/4dhj3v7y)
The reported results are averaged over 5 different runs.
%
For NBFNet, we used the same hyperparameter values as \citet{DBLP:conf/nips/ZhuZXT21} for WN18RR. For FB15k-237, we also used the same values, except that remove\_one\_hop was set to False, as this provides the best results. For NELL-995, we used the same hyperparameter values as for FB15k-237. 



\subsection{Results}
Tables \ref{tabMainFB}--\ref{tabMainNELL} show the results for the full evaluation setting, where all entities from the test graph are considered as candidates. In Table \ref{tabHITS50} we show the results of the reduced setting, with 50 randomly sampled negatives. As can be seen in Table \ref{tabHITS50}, the results for this latter setting are less informative, with many of our hybrid strategies achieving nearly identical results. This is due to the fact that Hits@10 is a very loose evaluation metric when only 50 negatives are considered. We include these results to allow for a comparison with published results of baselines such as GraIL. For the remainder of this section, we will focus on the full evaluation setting.

\smallskip
\noindent\textbf{Reranking the entities in $\mathcal{A}_q$ is helpful.} 
Comparing AnyBURL with the the GNN strategies (R-GCN and CompGCN), we can see that the latter overall perform better. The clearest improvements can be seen on FB15k-237 v3 and v4, WN18RR (all versions), and NELL-995 v1. These are the datasets with the highest amount of training data per relation, i.e.\ they either have large training sets or a small number of relations. This means that a larger number of high-quality rules per relation can typically be learned, which makes it more likely that multiple relevant rules can be found for a given prediction. For the same reason, we can also see that Noisy-or performs well on these datasets. Noisy-or overall achieves surprisingly strong results for H@10, where it often outperforms the GNN-based models. For the other metrics, the GNN-based models are better, but Noisy-or often still manages to outperform AnyBURL.
In the case of NELL-995, the GNN models are less successful, performing slightly better than AnyBURL on v1 and v4 but clearly worse on v2 and v3. NELL-995 is far noisier than the other KGs, having been extracted from text, which seems to affect the GNN-based models.

\smallskip
\noindent\textbf{Reranking the entities in $\mathcal{B}_q$ is helpful.}
We now look at the impact of reranking the entities in $\mathcal{B}_q$ using NBFNet, which we can see, e.g., by comparing AnyBURL + NBFNet with AnyBURL. We find that reranking the entities in $\mathcal{B}_q$ consistently improves the results, especially for FB15k-237 and NELL-995. For the WN18RR datasets, however, the impact of this reranking step is marginal.

\smallskip
\noindent\textbf{L1 and L2 largely explain the underperformance of AnyBURL.}
The R-GCN + NBFNet and CompGCN + NBFNet strategies were specfically designed to address limitations L1 and L2 of AnyBURL. These models largely keep the interpretability advantage of rule-base methods, at least when it comes to how the entities in $\mathcal{A}_q$ are ranked, given that our rule instantiation graphs cover at most five rules. Moreover, their performance is close to that of NBFNet, which supports the hypothesis that L1 and L2 are mostly responsible for the underperformance of AnyBURL. NBFNet generally still outperforms R-GCN + NBFNet and CompGCN + NBFNet, but the latter methods also outperform NBFNet in a few cases.

\smallskip
\noindent\textbf{NBFNet + NBFNet outperforms NBFNet.}
The only exceptions are  H@3 for FB15k-237 v1,  H@10 for WN18RR v1 and H@3 for NELL-995 v1. This supports the idea that NBFNet is prone to pick up spurious correlations, which rule-based methods can help to address. Note that the outperformance over NBFNet is particularly pronounced for the NELL-995 datasets. These datasets are much noisier, making it more likely for NBFNet to learn such spurious correlations. As mentioned before, the GNN-based rule aggregation strategies struggle on the NELL-995 datasets for a similar reason.
From a practical point of view, NBFNet + NBFNet is essentially as efficient as NBFNet, since the overhead of running AnyBURL is negligible. This strategy thus offers an effective and easy-to-use approach for improving state-of-the-art GNN models. 

\begin{figure}
\includegraphics[width=220pt]{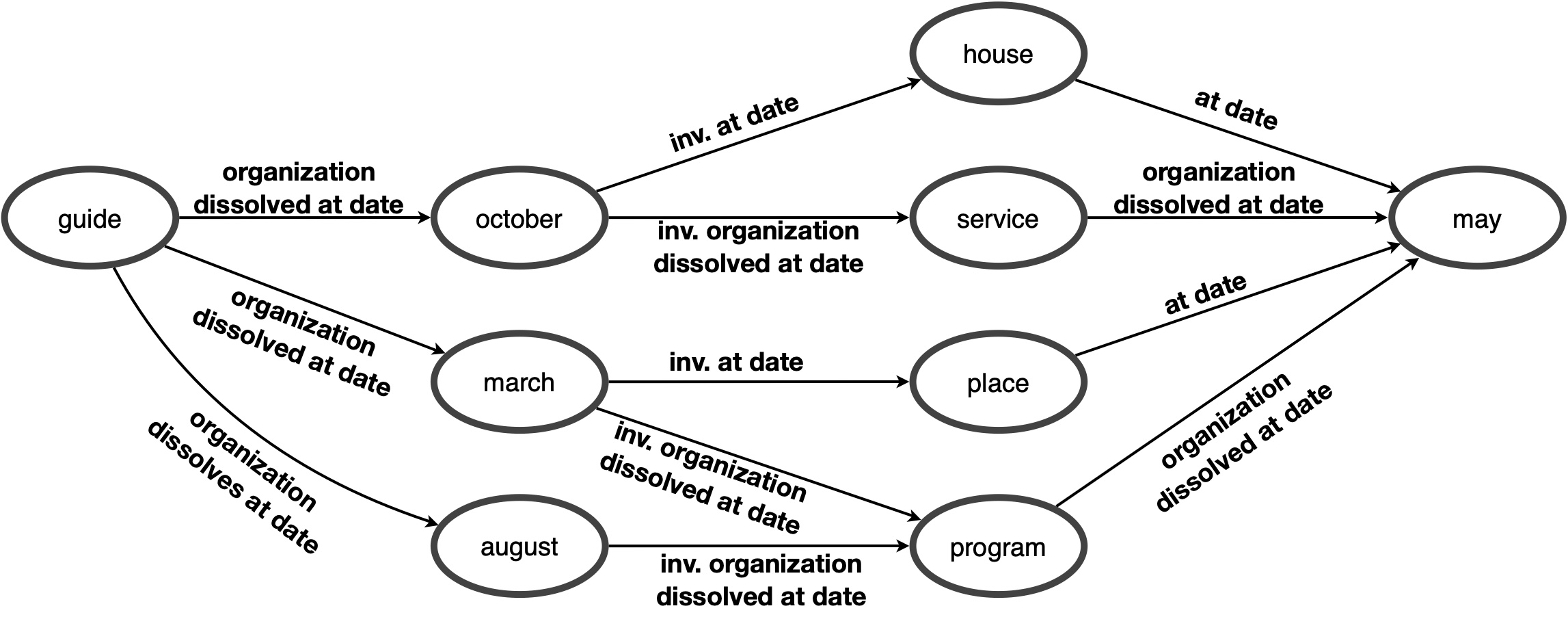}
\caption{NELL-995 v2 (\emph{guide}, 
\emph{organization dissolved at date}, \emph{may}).
\label{figNELL2Example}}.
\end{figure}

\begin{figure}
\includegraphics[width=200pt]{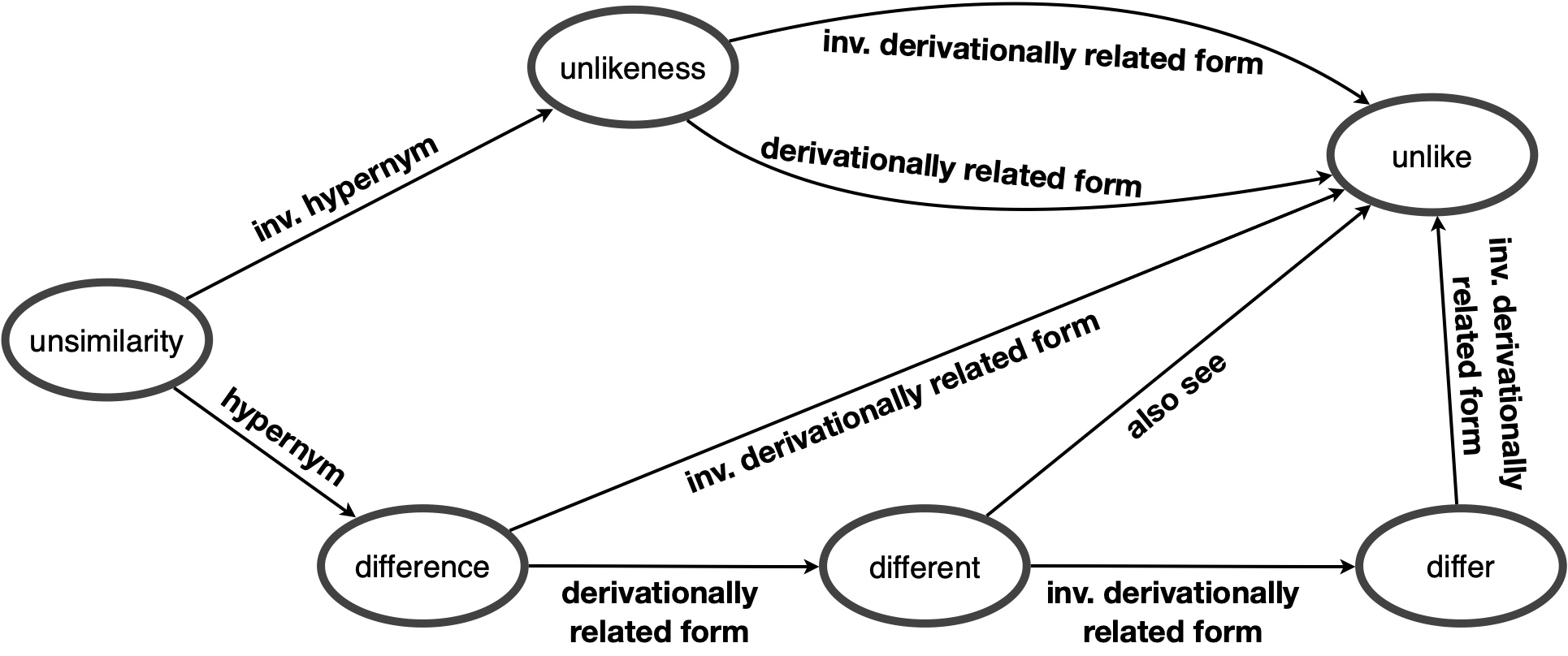}
\caption{WN18RR v1 (\emph{unsimilarity}, 
\emph{derivationally related form}, \emph{unlike}).
\label{figWNExample}}


\end{figure}

\begin{figure}
\centering
\includegraphics[width=200pt]{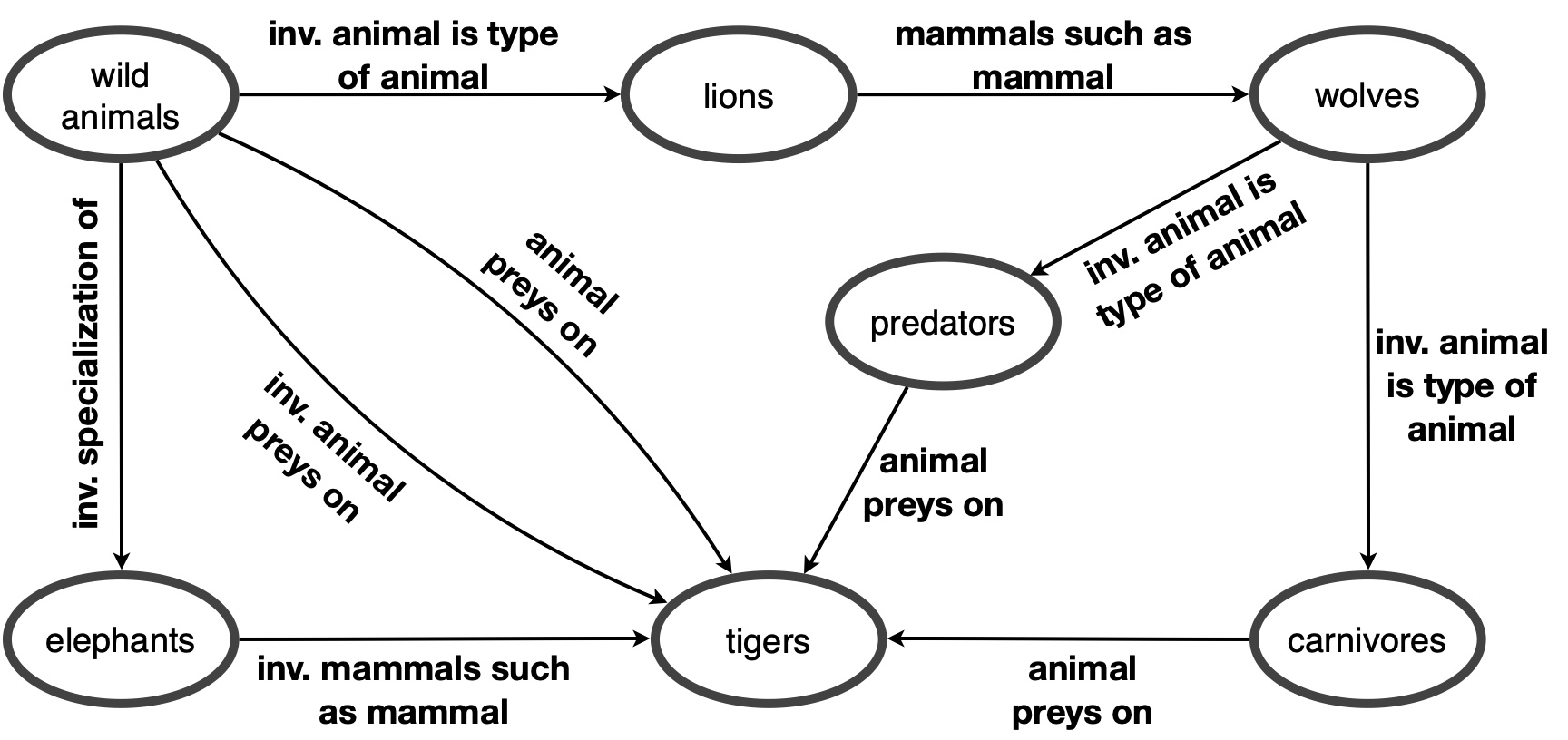}
\caption{NELL-995 v3 (\emph{wild animals}, 
\emph{inv.\ animal is type of animal}, \emph{tigers}).\label{figNELL3Example}}
\end{figure}

\subsection{Analysis}

\smallskip \noindent \textbf{Rare vs frequent relations.}
We found that the GNN-based aggregation methods can underperform when insufficient training data is available. A similar issue arises for relations that are too rare, even when the overall training graph is larger. To illustrate this, Figure \ref{figCommonRare} shows a break-down of the results (on the test set) for the 10\% most frequent relations in the training graph
(\emph{frequent}) the next 40\% most frequent relations (\emph{common}) and the 50\% least frequent relations (\emph{rare}). We can see that the outperformance of CompGCN over AnyBURL is generally far more pronounced for frequent relations. This is as expected, since we need sufficient training data to learn the GNN models. However, for NELL-995 v2 and v3, we see the opposite behaviour, with CompGCN clearly underperforming AnyBURL for frequent relations. This discrepancy arises from the noisy nature of these datasets, which means that a high number of unreliable rules are being learned. A particularly clear-cut example is provided in Figure \ref{figNELL2Example}, which shows a rule instantiation graph with rules that are essentially non-sensical. As another observation in Figure \ref{figCommonRare}, we can see that the outperformance of CompGCN + NBFNet over CompGCN is most pronounced for rare relations (especially for the FB15k-237 and NELL-995 datasets). This is also as expected, since we have fewer rules for rare relations, making it more likely that the correct answer ends up in $\mathcal{B}_q$.

\smallskip
\noindent\textbf{Rules offer weak evidence for WN18RR.}
For the WN18RR datasets, we found that the GNN-based rule aggregation strategies were able to outperform AnyBURL by a considerable margin. This reflects the fact that the available rules for these datasets offer relatively weak evidence. It therefore becomes more important to aggregate the evidence from multiple rule instantiations. The weak nature of the individual rules is illustrated in Figure \ref{figWNExample}, where we show a rule instantiation graph from WN18RR v1. Each of the rules individually only gives weak evidence, but the information encoded in the full rule instantiation graph nonetheless strongly suggests that there is a link between the two entities.

\smallskip
\noindent\textbf{Rule aggregation is less successful for NELL-995.}
As we already mentioned, NELL-995 is much noisier than the other KGs, due to the fact that this dataset was extracted from text. Figure \ref{figNELL3Example} illustrates this with an example from NELL-995 v3. In this case, most of the paths in the rule instantiation graph are not actually predictive (e.g.\ the fact that wild animals prey on tigers does not imply that tigers are a kind of wild animal). However, there are also rules which are somewhat more informative. For instance, the path in the bottom left corner expresses that tigers have something in common with elephants and that elephants are wild animals; this indeed provides us with some evidence for the inference that tigers could be wild animals. Due to the noisy nature of such rule instantiation graphs, focusing on the most reliable evidence (as done by AnyBURL) leads to better results than taking into account the entire rule instantiation graph.

\section{Conclusions}
We have analysed the performance gap between rule-based methods and GNNs for inductive KG completion. One important finding is that by reranking the entities which receive zero confidence by AnyBURL, we can significantly improve the performance of this rule-based method. We have also proposed to use a GNN to rerank the candidates predicted by AnyBURL. Rather than using the full knowledge graph, our GNN models only see the ground rules that predict each answer candidate. As such, their sole purpose is to aggregate the evidence provided by different AnyBURL rules. We found this strategy to be highly effective on several datasets. Together, these two modifications enable results which are close to those of NBFNet, while largely keeping the interpretability advantage of rule-based methods. We also analysed a variant in which NBFNet was used to rerank the candidates predicted by AnyBURL. This simple strategy allowed us to consistently outperform NBFNet. Finally, we have uncovered important methodological differences in how different inductive KG completion methods have been evaluated, meaning that published results are often incomparable.

\section{Acknowledgements}
This research was supported by the Leverhulme Trust (project RPG-2021-140) and undertaken using the supercomputing facilities at Cardiff University operated by Advanced Research Computing at Cardiff (ARCCA) on behalf of the Cardiff
Supercomputing Facility and the HPC Wales and
Supercomputing Wales (SCW) projects. We acknowledge the support of the latter, which is part funded by the European Regional Development Fund (ERDF) via the Welsh Government.

\section{Bibliographical References}\label{sec:reference}

\bibliographystyle{lrec-coling2024-natbib}
\bibliography{refs}

\appendix
\section{Appendix}

\noindent \textbf{Datasets Statistics}. Some statistics of the datasets are shown in Table \ref{tabDatasetStatistics}. Apart from the number of relations, entities and triples in the training and test graphs, we also report some statistics which depend on the AnyBURL rules. First, we report the total number of rules that were learned (\#Rules). We furthermore report statistics on the size of $\mathcal{A}_q$. In particular, we show how often $|\mathcal{A}_q|=0$, which is important because using AnyBURL makes no difference in those cases. We also show how often $|\mathcal{A}_q|=1$, which is important because then it does not matter which strategy is used for ranking the entities in $\mathcal{A}_q$. Next, we show how often $|\mathcal{A}_q|>10$, because in those cases reranking the entities in $\mathcal{B}_q$ does not affect the Hits@10 evaluation metric. Finally, we also report the average number of rule instantiations we have for the correct answer (Rule inst.).

\begin{table*}[]
\centering
\caption{Statistics for the considered datasets, showing the number of relations in the training ($|\mathcal{R}_{\textit{train}}|$) and test graph ($|\mathcal{R}_{\textit{test}}|$), the number of entities in the training ($|\mathcal{E}_{\textit{train}}|$) and test graph ($|\mathcal{E}_{\textit{test}}|$), the number of triples in the training ($|\mathcal{G}_{\textit{train}}|$) and test graph ($|\mathcal{G}_{\textit{test}}|$), the number of rules learned by AnyBURL (\#Rules), the percentage of link prediction queries in the test set for which $\mathcal{A}_q$ contains 0, 1, and more 10 entities, and the average number of rule instantiations for the correct answer (Rule inst.).\label{tabDatasetStatistics}}
\setlength{\tabcolsep}{3pt}
\begin{tabular}{ll ccc ccc ccccc}
\toprule
& & $|\mathcal{R}_{\textit{train}}|$ & $|\mathcal{E}_{\textit{train}}|$ & $|\mathcal{G}_{\textit{train}}|$ & $|\mathcal{R}_{\textit{test}}|$ & $|\mathcal{E}_{\textit{test}}|$ & $|\mathcal{G}_{\textit{test}}|$ & \#Rules & $|\mathcal{A}_q|= 0$ & $|\mathcal{A}_q| = 1$  & $|\mathcal{A}_q|> 10$ & Rule inst.\\
\midrule
\parbox[t]{2mm}{\multirow{4}{*}{\rotatebox[origin=c]{90}{\footnotesize\textbf{FB15k-237}}}} & v1 & 180 & 1594 & 5226 & 142 & 1093 & 2404 & 10575&30.5&11.0&35.6&1.6 \\
& v2 & 200 & 2608 & 12085 & 172 & 1660 & 5092 & 19097&8.4&9.4&60.5&5.5 \\
& v3 & 215 & 3668 & 22394 & 183 & 2501 & 9137 & 18445&5.5&7.2&70.5&7.4 \\
& v4 & 219 & 4707 & 33916 & 200 & 3051 & 14554 & 16803&6.3&5.2&76.8&8.7 \\
\midrule
\parbox[t]{2mm}{\multirow{4}{*}{\rotatebox[origin=c]{90}{\footnotesize\textbf{WN18RR}}}} & v1 & 9 & 2746 & 6678 & 8 & 922 & 1991 & 1816&3.7&1.6&79.5&2.5 \\
& v2 & 10 & 6954 & 18968 & 10 & 2757 & 4863 & 2075&4.3&4.4&53.9&2.2 \\
& v3 & 11 & 12078 & 32150 & 11 & 5084 & 7470 & 2273&18.2&7.0&40.0&1.3 \\
& v4 & 9 & 3861 & 9842 & 9 & 7084 & 15157 & 2104&2.7&1.7&76.1&2.2 \\
\midrule
\parbox[t]{2mm}{\multirow{4}{*}{\rotatebox[origin=c]{90}{\footnotesize\textbf{NELL-995}}}} & v1 & 14 & 3103 & 5540 & 14 & 225 & 1034 & 1940&0.5&12.5&78.0&41.6 \\
& v2 & 88 & 2564 & 10109 & 79 & 2086 & 5521 & 6187&13.4&8.6&62.8& 10.0\\
& v3 & 142 & 4647 & 20117 & 122 & 3566 & 9668 & 8953&14.6&6.9&68.5&13.9 \\
& v4 & 76 & 2092 & 9289 & 61 & 2795 & 8520 & 6009&20.5&11.8&21.8&1.7 \\
\bottomrule
\end{tabular}
\end{table*}

\smallskip \noindent \textbf{AnyBURL Runtime}. 
AnyBURL is an anytime rule learner, which can learn rules for a given time duration. As we consider only a few rules with the highest confidence 
to construct  rule instantiation graphs, our hypothesis is that running AnyBURL for a short amount of time is sufficient. 
To test this hypothesis, we have analysed the effect of changing the runtime limit of AnyBURL.
We perform an ablation study on WN18RR (as it has fewer rules than other datasets) to assess the performance of CompGCN over the rule instantiation graphs.
Table~\ref{tab:abl_wn} shows the results of learning rules  for 10, 100, and 1000 seconds.
It is observed that for all  versions of WN18RR, the validation accuracy is comparable across  all time limits. 
Table~\ref{tab:abl_wn} also presents results of the performance of CompGCN with the above time limits. It is evident that CompGCN gets the best performance with a 10 seconds time limit.  

\begin{table*}
\centering
\caption{Effect of AnyBURL runtime limit when using CompGCN and the WN18RR datasets. H@10\_50 refers to Hits@10 with the reduced evaluation setting, where only 50 negatives are considered for each query.}
\label{tab:abl_wn}
\begin{tabular}{cccccccc}
\toprule
&                                                                          & \multicolumn{1}{c}{\textbf{Validation}} & \multicolumn{5}{c}{\textbf{Test}}  \\ 
\cmidrule(lr){3-3} \cmidrule(lr){4-8}
\multicolumn{1}{c}{\textbf{Version}}             & \textbf{AnyBURL Runtime} & \multicolumn{1}{c}{\textbf{Acc}}   & \multicolumn{1}{c}{\textbf{MRR}}  & \multicolumn{1}{c}{\textbf{H@1}}  & \multicolumn{1}{c}{\textbf{H@3}}  & \multicolumn{1}{c}{\textbf{H@10}} & \textbf{H@10\_50} \\ \midrule
\multicolumn{1}{c}{\multirow{3}{*}{v1}} & 10                                                         & \multicolumn{1}{c}{97.1}       & \multicolumn{1}{c}{68.1} & \multicolumn{1}{c}{62.0}   & \multicolumn{1}{c}{72.1} & \multicolumn{1}{c}{78.4} & 86.4     \\ 
\multicolumn{1}{c}{}                    & 100                                                        & \multicolumn{1}{c}{97.1}       & \multicolumn{1}{c}{64.8} & \multicolumn{1}{c}{58.5} & \multicolumn{1}{c}{68.4} & \multicolumn{1}{c}{75.9} & 84.8     \\ 
\multicolumn{1}{c}{}                    & 1000                                                       & \multicolumn{1}{c}{97.0}         & \multicolumn{1}{c}{62.6} & \multicolumn{1}{c}{55.3} & \multicolumn{1}{c}{67.1} & \multicolumn{1}{c}{75.7} & 84.9     \\ \midrule
\multicolumn{1}{c}{\multirow{3}{*}{v2}} & 10                                                         & \multicolumn{1}{c}{96.5}       & \multicolumn{1}{c}{65.3} & \multicolumn{1}{c}{59.0}   & \multicolumn{1}{c}{69.9} & \multicolumn{1}{c}{76.3} & 82.5     \\ 
\multicolumn{1}{c}{}                    & 100                                                        & \multicolumn{1}{c}{96.7}       & \multicolumn{1}{c}{62.6} & \multicolumn{1}{c}{56.0}   & \multicolumn{1}{c}{66.9} & \multicolumn{1}{c}{75.2} & 82.2     \\ 
\multicolumn{1}{c}{}                    & 1000                                                       & \multicolumn{1}{c}{96.7}       & \multicolumn{1}{c}{63.8} & \multicolumn{1}{c}{57.0}   & \multicolumn{1}{c}{68.3} & \multicolumn{1}{c}{76.1} & 83.0       \\ \midrule
\multicolumn{1}{c}{\multirow{3}{*}{v3}} & 10                                                         & \multicolumn{1}{c}{96.9}       & \multicolumn{1}{c}{40.7} & \multicolumn{1}{c}{36.0}   & \multicolumn{1}{c}{43.4} & \multicolumn{1}{c}{48.5} & 65.5     \\ 
\multicolumn{1}{c}{}                    & 100                                                        & \multicolumn{1}{c}{97.0}         & \multicolumn{1}{c}{40.0}   & \multicolumn{1}{c}{35.4} & \multicolumn{1}{c}{42.9} & \multicolumn{1}{c}{47.6} & 65.2     \\ 
\multicolumn{1}{c}{}                    & 1000                                                       & \multicolumn{1}{c}{97.0}         & \multicolumn{1}{c}{39.6} & \multicolumn{1}{c}{35.3} & \multicolumn{1}{c}{42.4} & \multicolumn{1}{c}{47.2} & 64.9     \\ \midrule
\multicolumn{1}{c}{\multirow{3}{*}{v4}} & 10                                                         & \multicolumn{1}{c}{97.0}         & \multicolumn{1}{c}{60.3} & \multicolumn{1}{c}{54.5} & \multicolumn{1}{c}{64.1} & \multicolumn{1}{c}{70.1} & 79.7     \\ 
\multicolumn{1}{c}{}                    & 100                                                        & \multicolumn{1}{c}{97.1}       & \multicolumn{1}{c}{58.3} & \multicolumn{1}{c}{53.0}   & \multicolumn{1}{c}{61.4} & \multicolumn{1}{c}{68.0}   & 78.4     \\ 
\multicolumn{1}{c}{}                    & 1000                                                       & \multicolumn{1}{c}{97.0}         & \multicolumn{1}{c}{58.5} & \multicolumn{1}{c}{52.9} & \multicolumn{1}{c}{61.9} & \multicolumn{1}{c}{68.0}   & 78.5     \\ \bottomrule
\end{tabular}
\end{table*}


\smallskip \noindent \textbf{Selecting Top k Rules}. When constructing a rule instantiation graph, we use $k$ rules with highest confidence. Requiring only a small value of $k$ is preferred  as it preserves the interpretability. To find a suitable $k$, we further conduct an ablation study which looks at the performance of CompGCN when aggregating rule instantiation graphs constructed using  $k$ rules, with $k \in \{5, 10, 50, 100,  1000\}$. Table~\ref{tab:ablation_top_k} shows that the validation accuracy is comparable across all  $k$ values for all  versions of FB15k-237. Further, it shows that CompGCN performs comparably across all $k$ values. 


\begin{table*}[t]
\centering
\caption{Effect of varying the number of top-$k$ rules when using CompGCN KG and the FB15k-237 datasets. H@10\_50 refers to Hits@10 with the reduced evaluation setting, where only 50 negatives are considered for each query.}
\label{tab:ablation_top_k}
\begin{tabular}{cccccccc}
\toprule
\multicolumn{2}{l}{}                           & \multicolumn{1}{c}{\textbf{Validation}} & \multicolumn{5}{c}{\textbf{Test}}                                                                                                                  \\ 
\cmidrule(lr){3-3}\cmidrule(lr){4-8}
\multicolumn{1}{c}{\textbf{Version}}    & \textbf{Top $k$ Rules}   & \multicolumn{1}{c}{\textbf{Acc}}   & \multicolumn{1}{c}{\textbf{MRR}} & \multicolumn{1}{c}{\textbf{H@1}} & \multicolumn{1}{c}{\textbf{H@3}} & \multicolumn{1}{c}{\textbf{H@10}} & \multicolumn{1}{c}{\textbf{H@10\_50}} \\ \midrule
\multirow{5}{*}{v1}              & 5             & \multicolumn{1}{c}{90.8}       & 37.0                       & 31.4                     & 41.3                     & 45.6                      & 60.4                          \\ 
                                 & 10            & \multicolumn{1}{c}{90.8}       & 37.4                     & 31.6                     & 41.6                     & 45.4                      & 60.2                          \\ 
                                 & 50            & \multicolumn{1}{c}{90.8}       & 35.6                     & 30.3                     & 39.0                       & 42.6                      & 58.3                          \\ 
                                 & 100           & \multicolumn{1}{c}{89.8}       & 38.0                       & 32.5                     & 41.8                     & 45.3                      & 61.0                            \\ 
                                 & 1000          & \multicolumn{1}{c}{90.0}         & 37.1                     & 31.6                     & 41.0                       & 44.9                      & 60.4                          \\ \midrule 
\multirow{5}{*}{v2}              & 5             & \multicolumn{1}{c}{89.9}       & 45.1                     & 36.2                     & 50.3                     & 60.4                      & 82.0                            \\ 
                                 & 10            & \multicolumn{1}{c}{89.9}       & 44.2                     & 35.7                     & 49.0                       & 59.4                      & 81.2                          \\ 
                                 & 50            & \multicolumn{1}{c}{90.0}         & 44.5                     & 35.6                     & 50.9                     & 59.7                      & 81.4                          \\ 
                                 & 100           & \multicolumn{1}{c}{89.8}       & 44.4                     & 35.4                     & 50.9                     & 60.8                      & 81.4                          \\ 
                                 & 1000          & \multicolumn{1}{c}{89.4}       & 44.1                     & 34.5                     & 50.3                     & 60.4                      & 81.4                          \\ \midrule 
\multirow{5}{*}{v3}              & 5             & \multicolumn{1}{c}{89.6}       & 40.7                     & 31.5                     & 47.3                     & 55.8                      & 83.1                          \\ 
                                 & 10            & \multicolumn{1}{c}{89.6}       & 36.8                     & 28.1                     & 42.5                     & 51.3                      & 82.3                          \\ 
                                 & 50            & \multicolumn{1}{c}{89.2}       & 39.6                     & 31.3                     & 45.3                     & 53.2                      & 81.8                          \\ 
                                 & 100           & \multicolumn{1}{c}{88.9}       & 39.6                     & 31.4                     & 44.7                     & 53.1                      & 82.4                          \\ 
                                 & 1000          & \multicolumn{1}{c}{88.8}       & 38.0                       & 29.4                     & 43.4                     & 52.8                      & 82.7                          \\ \midrule 
\multirow{5}{*}{v4}              & 5             & \multicolumn{1}{c}{88.9}       & 38.3                     & 29.5                     & 43.4                     & 53.3                      & 84.7                          \\ 
                                 & 10            & \multicolumn{1}{c}{88.8}       & 37.0                       & 28.2                     & 41.9                     & 51.8                      & 84.0                            \\ 
                                 & 50            & \multicolumn{1}{c}{88.6}       & 35.6                     & 27.1                     & 40.6                     & 51.1                      & 83.8                          \\ 
                                 & 100           & \multicolumn{1}{c}{88.6}       & 37.5                     & 28.5                     & 43.0                       & 53.2                      & 84.0                            \\ 
                                 & 1000          & \multicolumn{1}{c}{88.0}         & 38.1                     & 29.7                     & 42.8                     & 53.1                      & 84.3                          \\ \bottomrule 
\end{tabular}
\end{table*}


\end{document}